\documentclass[10pt,a4paper]{book}

\usepackage{sty-cividale-procs-WS}

\begin{document}

\listauthors{M.~Frailis, O.~Mansutti,
P.~Boinee, G.~Cabras, A.~De Angelis, B.~De Lotto, M.~Dell'Orso, A.~Forti, M.~Mariotti, R.~Paoletti, L.~Peruzzo, A.~Saggion, A.~Scribano, N.~Turini}
\index{De Angelis, A.} \index{Frailis, M.} \index{Mansutti, O.} \index{Boinee, P.} \index{Cabras, G.} \index{De Lotto, B.} \index{Forti, A.} \index{Dell'Orso, M.} \index{Scribano, A.} \index{Preuzzo, L.} \index{Mariotti, M.} \index{Saggion, A.} \index{Paoletti, R.} \index{Turini, N.}

\talkauthor{Marco Frailis, Oriana Mansutti,\\
Praveen Boinee, Giuseppe Cabras, Alessandro De Angelis, Barbara De Lotto, Alberto Forti}
\talkaddress{Dipartimento di Fisica, Universit\`a di Udine \\%
         \& INFN Sezione di Trieste - Gruppo collegato di Udine, \\%
         Via delle Scienze 208 -- 33100 Udine, Italy \\%
         E-mail: deangelis@fisica.uniud.it, frailis@fisica.uniud.it, mansutti@fisica.uniud.it}

\talkauthor{Mauro Dell'Orso}
\talkaddress{Universit\`a di Pisa\\Lungarno Pacinotti, 43 -- 56126 Pisa, Italy}

\talkauthor{Riccardo Paoletti, Angelo Scribano, Nicola Turini}
\talkaddress{Universit\`a di Siena\\via Banchi di Sotto, 55 -- 53100 Siena, Italy}

\talkauthor{Mosè Mariotti, Luigi Peruzzo, Antonio Saggion}
\talkaddress{Universit\`a di Padova\\via 8 Febbraio, 2 -- 35122 Padova, Italy}

\talktitle{A third level trigger programmable on FPGA for the gamma/hadron separation in a Cherenkov telescope using pseudo-Zernike moments and the SVM classifier}{A third level trigger programmable on FPGA for the gamma/hadron separation in a Cherenkov telescope using pseudo-Zernike moments and the SVM classifier}


%


%
%
%

\abstracts{
We studied the application of the Pseudo-Zernike features as image parameters (instead of the Hillas parameters) for the  discrimination between the images produced by atmospheric electromagnetic showers caused by gamma-rays and the ones produced by atmospheric electromagnetic showers caused by hadrons in the MAGIC Experiment.
We used a Support Vector Machine as classification algorithm with the computed Pseudo-Zernike features as classification parameters.
We implemented on a FPGA board a kernel function of the SVM and the Pseudo-Zernike features to build a third level trigger for the gamma-hadron separation task of the MAGIC Experiment.
}

\section{Introduction}

Gamma-ray detectors experiments have the need of separating signals produced by photons from signals produced by hadrons.

In the case of the MAGIC telescope, the need is to discriminate between the images produced by atmospheric electromagnetic showers caused by gamma-rays and the ones produced by atmospheric electromagnetic showers caused by hadrons.

To contrubute to this discrimination, we studied some approaches applicable to the gamma-hadron separation of the MAGIC telescope experiment.

We set up our work as a two-class image classification problem.
The novelties we propose lie in two phases of the consutruction of a classifier for the MAGIC images: in the choice of the classification parameters and in the choice of the classification algorithm.

We used the results of these classification problem to build a third level trigger for the gamma-hadron separation task of the MAGIC Experiment by implementing a kernel function of the classificator and the pseudo-Zernike featrues on a FPGA board, which can be used to discriminate the images coming from the MAGIC telescope.

\section{Classification Parameters}

The image parameters usually used for image classification and analysis in the MAGIC telescope collaboration are the Hillas parameters~\cite{mainHillas,HillasWittek}.

In our work we considered the application of the pseudo-Zernike features~\cite{mainPseudoZernike} as image parameters for the MAGIC telescope images.

The \emph{Zernike moments}~\cite{mainZernike} are an infinite orthogonal basis of polynomials for the representation of image functions. The value of the coefficient of a basis element is referred to as \emph{Zernike feature} of the image.

A first characteristic of this basis that makes it of interest for the MAGIC telescope images is the rotation invariance of the Zernike moments, that makes the representation of the image independent from a rotation of the telescope along its symmetry axis.

Two other characteristics of Zernike moments that are important for the MAGIC telescope images are that they are not scale and translation invariant (though they can be modified to get these invariances):
\begin{itemize}
\item avoiding the scale invariance makes these parameters sensitive to the magnitude of the image, and this means that the further classification will take into account the energy of the originating particle (and the classifier automatically behaves differently for low energy and high energy images);
\item avoiding the translation invariance makes these parameters sensitive to the Hillas alpha parameter of the image (the transformation that leads to translation invariance would make all alphas equal to zero).
\end{itemize}

The orthogonality property of Zernike moments enables one to select the required number of features to get a good enough representation of the image, since orthogonality:
\begin{itemize}
\item makes the image reconstruction from its features computationally simple;
\item enables one to evaluate the image representation ability (contribution to the reconstruction process) of each order moment.
\end{itemize}

Unfortunately, high order Zernike moments are very sensitive to noise~\cite{noiseZernike}. 
In order to solve this problem and keep all the useful characteristics of Zernike moments, we have considered the pseudo-Zernike moments.

Like the Zernike moments, also the \emph{pseudo-Zernike moments}~\cite{mainPseudoZernike,tesiMazza} are an infinite orthogonal basis of polynomials for the representation of image functions (where the value of the coefficient of a basis element is referred to as \emph{pseudo-Zernike feature} of the image), and are not scale and tanslation invariant.

Pseudo Zernike moments are more robust to image noise, since the number of pseudo-Zernike polinomials of a fixed maximum order $n$ is $(n+1)^2$, while the Zernike moments are in number of $\begin{aligned}\frac{1}{2}(n+1)(n+2)\end{aligned}$~\cite{computePseudoZernike}.


\section{Classification Algorithm and Training Strategy}

Both with the choice of the Hillas parameters and the pseudo-Zernike features as classification parameters, a classification algorithm has to be chosen (for which the selected parameters are the input).

The classification algorithm we used is based on the usage of a Support Vector Machine (SVM) as classification technique~\cite{svmBurges}, since they are currently under very active research within the fields of machine and statistical learning and they had not been conclusively tested for our classification task~\cite{svmBock}. The SVM classification method is systematic, reproducible, and properly grounded by  the statistical learning theory.

The SVM algorithm solves an optimization problem on a set of training data to determine the parameters of a function (the \emph{decision function}) to be evaluated on the data that have to be classified.
The data have to be described through several features (in our case the Hillas parameters or the pseudo-Zernike features) and the evaluation of the decision function must give a target value that represents the result of the classification task (in our case, a value telling if the the image was produced by a primaty photon or hadron, that can be represented by a value in $\{1,-1\}$)~\cite{svmHsu,svmBen}.

In its simplest form, an SVM is able to perform a binary classification finding the ``best'' separating hyperplane between two linearly separable classes. There are infinite hyperplanes properly separating the data. So, the SVM finds this hyperplane maximizing the distance, or \emph{margin}, between the \emph{support hyperplanes} for each class. A hyperplane supports a class if all points in that class are on one side of that plane. This problem is formulated as a quadratic programming problem (QP) and can be solved by effective robust algorithms. If the data is not linearly separable, slack variables are introduced into the QP problem to accept outliers.

In order to learn non-linear relations, SVM implicitly applies a fixed non-linear mapping of the data to a enough high (maybe infinite) dimensional feature space in which the classification through the hyperplanes can be used~\cite{svmHsu}. The implicit mapping is performed by using the so called \emph{kernel functions}, that one can choose according to which function makes the SVM perform better~\cite{svmBen}. In the leterature, several kernel functions have been proposed~\cite{svmHsu}. Among them, the Gaussian radial basis function kernel is tipically used for classification tasks. It is defined by:
$$
k(x,y)=e^{-\gamma\|x-z\|^2}
$$
where $x$ and $z$ are input data and $\gamma$ is a parameter specified by the user regulating the width of the Gaussian kernel.



\section{Training the SVM algorithm}

The SVM algorithm with the Gaussian kernel requires two parameters: the parameter $C>0$ is a regularization constant determining a trade-off between the empirical error (number of wrongly classified inputs) and the \emph{complexity} of the found solution; the $\gamma$ parameter of the Gaussian kernel affects the complexity of the decision boundary.

To select the best $C$ and $\gamma$ parameters for the training step, we used a cross validation via grid search. With this method, the training set is first separated into $m$ folds. Sequentially a fold is considered as the validation set and the rest are for training. So an adaptive grid-search is performed on $C$ and $\gamma$, trying exponentially growing sequences for the two parameters.

\section{Case study: gamma/hadron separation in MAGIC}

To test both the pseudo-Zernike features and the SVM classification algorithm for the gamma/hadron separation problem in a Cherenkov telescope we used the software infrastructure built for the MAGIC experiment.

First, a dataset of simulated gamma-ray photons has been produced, generating calibrated Cherenkov images for each event, with different pointing positions.
Then, real OFF data (i.e.~data taken from a sky region not too far away from the Crab source) gathered by the MAGIC telescope has been used to represent the background (hadrons, electrons, muons, diffuse photons).

From these dataset we extracted a training set of 12228 gammas and 12306 hadrons and a test set of 6109 gammas and 6183 hadrons (using a random sampling).
We considered events taken with pointing position comprised between 0 and 12 degrees.

We modified the MAGIC software pipeline to extract the Zernike features from each Cherenkov image but using the standard image cleaning method currently adopted in MAGIC to select only the pixels above the pedestal signal.

After extracting the pseudo-Zernike features from each image (we use an order $n=7$, corresponding to 36 features), we normalize the training and test set with respect to the mean and standard deviation of each feature (calculated on the training set).
This is performed to avoid features in greater numeric ranges dominate those in smaller numeric ranges. 
With this method, each feature varies approximately in the range $[-1, 1]$.

\begin{figure}[ht]
\centering \vspace*{-6.5mm}
\includegraphics[width=\textwidth]{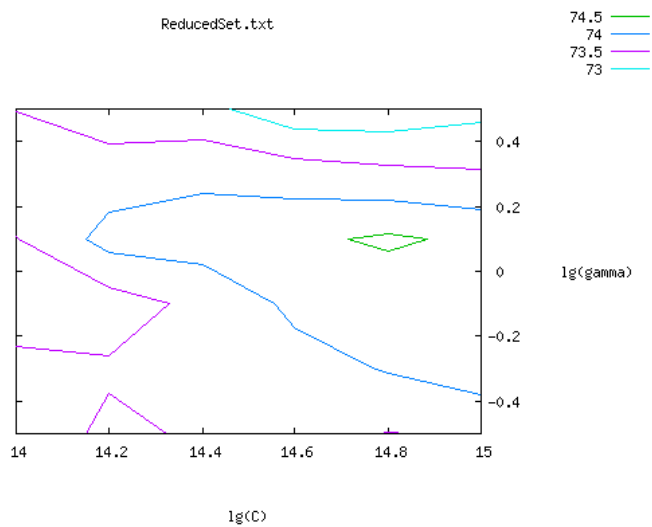} \vspace*{-9.4mm}
\caption{Cross validation with grid search.\label{frailis-fig.1}}
\end{figure}

After applying the cross validation via grid search (see Figure~\ref{frailis-fig.1}) on a small randomly sampled subset of the training set (5\% of the entire set), we have obtained the best rates for $C=28526.2$ and $\gamma=1.07$.
Finally, we trained the SVM classifier with the aforementioned parameters obtaining the following results on the test set:
\begin{center}
{\footnotesize
\begin{tabular}{|c|c|c|c|}
\hline
        & Total & Recognized & Ratio  \\
\hline
 Gammas &  6109 &       5271 & 86.3\% \\
\hline
Hadrons &  6183 &       4259 & 68.9\% \\
\hline
\end{tabular}
}
\end{center}
The overall accuracy obtained is 77.5\%.

\section{FPGA implementation}

A Xilinx Spartan 3 FPGA board was chosen to implement both the feature extractor (computing the pseudo-Zernike moments) and the SVM decision function while the training algorithm is meant to be performed off-line via software. Alternative FPGA-based implementations of Zernike moments and Support Vector Machines can be found in \cite{fpgaZer,fpgaSVM}.

\begin{figure}[ht]
\centering
\includegraphics[width=0.5\textwidth]{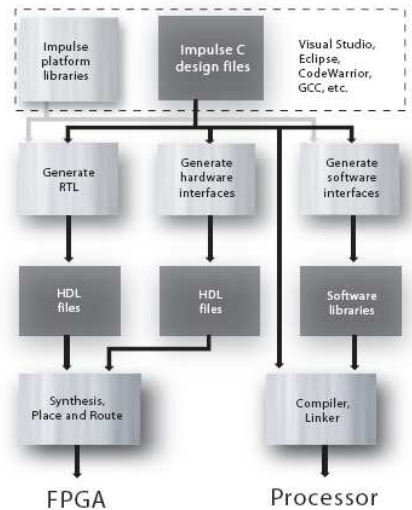}
\caption{C to HDL conversion with ImpulseC.\label{frailis-fig.2}}
\end{figure}

As programming language we chose a commercial product, ImpulseC, a C to RTL/HDL compiler (see Figure~\ref{frailis-fig.2}). 
With this approach we can reason in terms of algorithms, not of the hardware logic, requiring less efforts to convert the original C program to the FPGA device~\cite{fpgaProg}.

Currently, ImpulseC does not support all the features of the C standard language for the hardware translation. 
In particular, it only supports integer arithmetic and fixed point basic arithmetic. 
No global variables or function calls can be used within the main function to be translated in VHDL.

Therefore, we had to implement from scratch the fixed point square root (to calculate zernike features) and the exponential function (used by the SVM decision function). 
To avoid function calls, we implemented them as C macros. 
To improve the efficiency of the pseudo-Zernike features computation and reduce the approximation errors (due to the fixed point arithmetic), the pseudo-Zernike radial polynomials, which depend only on the pixel coordinates of the telescope, are calculated only once via software and stored as fixed point values in the FPGA.




\end{document}